\documentclass[letterpaper,10pt,conference]{ieeeconf}  
\IEEEoverridecommandlockouts                              
\usepackage{subcaption}
\usepackage{hyperref}
\usepackage{url}
\usepackage{cite}
\usepackage{amsmath,amssymb,amsfonts}
\usepackage{bbm}
\usepackage{algorithmic}
\usepackage{graphicx}
\usepackage{textcomp}
\usepackage[table]{xcolor}
\usepackage{mwe}
\usepackage{multirow}
\usepackage{hhline}
\usepackage{pifont} 

\def\BibTeX{{\rm B\kern-.05em{\sc i\kern-.025em b}\kern-.08em
    T\kern-.1667em\lower.7ex\hbox{E}\kern-.125emX}}
    
\newcommand{\myParagraph}[1]{\noindent \textbf{#1} ---}

\newcommand{\etal}{{et al.}} 

\usepackage{colortbl}

\hypersetup{
    colorlinks=true,
    linkcolor=blue,
    filecolor=magenta,      
    urlcolor=cyan,
    pdftitle={Overleaf Example},
    pdfpagemode=FullScreen,
    }

\begin{document}

\title{An in-depth experimental study of sensor usage and \\ visual reasoning of robots navigating in real environments
\\
\thanks{We acknowledge support from
ANR through AI-chair grant ``Remember'' (ANR-20-CHIA-0018).}
}

\author{Assem Sadek$^{1,2}$, Guillaume Bono$^{1}$, Boris Chidlovskii$^{2}$ and Christian Wolf$^{1}$
\thanks{$^{1}$Assem Sadek, Guillaume Bono and Christian Wolf are with LIRIS, UMR CNRS 5205, Université de Lyon, INSA Lyon, Villeurbanne, France. {\tt\footnotesize firstname.lastname@insa-lyon.fr},}%
\thanks{$^{2}$Assem Sadek and Boris Chidlovskii are with Naver Labs Europe, ch. Maupertuis 6, Meylan, France. {\tt\footnotesize firstname.lastname@naverlabs.com}}
}

\maketitle

\begin{abstract}
Visual navigation by mobile robots is classically tackled through SLAM plus optimal planning, and more recently through end-to-end training of policies implemented as deep networks. While the former are often limited to waypoint planning, but have proven their efficiency even on real physical environments, the latter solutions are most frequently employed in simulation, but have been shown to be able learn more complex visual reasoning, involving complex semantical regularities.
Navigation by real robots in physical environments is still an open problem. End-to-end training approaches have been thoroughly tested in simulation only, with experiments involving real robots being restricted to rare performance evaluations in simplified laboratory conditions.  

In this work we present an in-depth study of the performance and reasoning capacities of \emph{real physical agents}, trained in simulation and deployed to two different physical environments. Beyond benchmarking, we provide insights into the generalization capabilities of different agents training in different conditions. We visualize sensor usage and the importance of the different types of signals. We show, that for the PointGoal task, an agent pre-trained on wide variety of tasks and fine-tuned on a simulated version of the target environment can reach competitive performance \emph{without modelling any sim2real transfer}, i.e. by deploying the trained agent directly from simulation to a real physical robot.
\end{abstract}

\section{Introduction}
\label{sec:introduction}
\noindent
The design of mobile robots capable of performing visual navigation tasks in real physical environments has been classically addressed with geometric pipelines creating maps from LIDAR or visual input, localizing themselves on these maps (often simultaneously --- SLAM), and using symbolic planners to navigate to a specified target position. In recent years, machine learning has had a deep impact on these problems either by extending the classical pipelines with semantic information of the scene, complementary to geometry, or through end-to-end learning of navigation policies directly mapping observations to actions and usually trained in simulation by Reinforcement learning (RL), Inverse RL, Imitation Learning, or a combination of objectives.

While classical tasks focus on waypoint navigation (PointGoal), which essentially requires the estimation of free navigational space and the computation of shortest paths, large-scale training in simulation has opened the door to more complex problems, which require more advanced visual and spatial reasoning. Examples are ObjectNav~\cite{ObjectNavRevisited}, which requires finding and recognizing objects, whereas Multi-ON \cite{DBLP:conf/nips/WaniPJCS20} and the K-item scenario \cite{Beeching2020ICPR,BeechingECMLPKDD2020} require mapping objects in some form of latent memory, allowing the agent to find them quickly when needed. This sub-field, Embodied Computer Vision, is heavily dominated by training \emph{and evaluation} in simulated photo-realistic 3D environments like Habitat \cite{Savva_2019_ICCV}, AI-Thor~\cite{ai2thor} etc. The transfer from simulation to real environments (``\emph{sim2real}'') is one of the main current challenges in robot learning and as such widely studied. However, thorough evaluations of navigation capabilities in real environments are rare, compared to other robotic tasks like grasping and object manipulations. Experiments with mobile robots are time-consuming, as they need to be monitored by human operators and the reproduction of identical or similar evaluation conditions is difficult.

\begin{figure}[t] \centering
{
\setlength{\tabcolsep}{1pt}
\begin{tabular}{cc}
\begin{minipage}{0.48\linewidth}
\includegraphics[width=\linewidth]{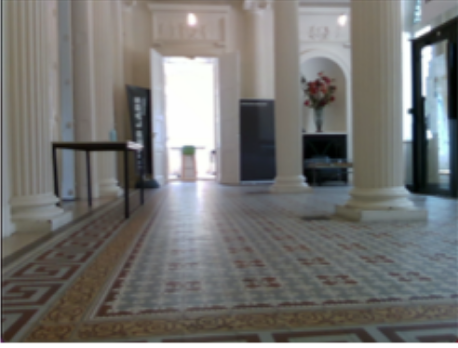}
\end{minipage}
&
\begin{minipage}{0.48\linewidth}
\includegraphics[width=\linewidth]{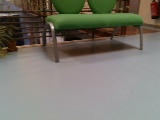}
\end{minipage}
\\
\begin{minipage}{0.48\linewidth}
\includegraphics[width=\linewidth]{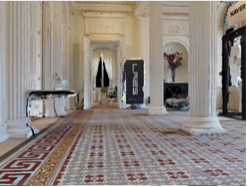}
\end{minipage}
&
\begin{minipage}{0.48\linewidth}
\includegraphics[width=\linewidth]{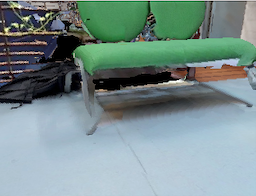}
\end{minipage}
\\
\end{tabular}
}
\caption{We present a deep experimental study of navigation capabilities of mobile robots in two different real physical (top) indoor environments: ``NLE'', a French 19th century 
furnished castle (left) and ``INSAL'', modern office spaces (right). The agents have been trained in different sets of simulated environments, which may contain, or not, 3D-scans of the evaluation environments (bottom: 2D observations from the simulator), targeting the evaluation of different generalization scenarios.
}
\label{fig:teaser}
\end{figure}

In this work we describe an in-depth study evaluating the performance of agents trained in simulation on two different physical environments. We chose the PointGoal task with GPS pointer estimated from a SLAM algorithm, which allows to compare the trained agents to classical algorithms based on SLAM and optimal planning. In our scenario, absolutely no adaptation is done for sim2real transfer: the agents are trained in simulation, and evaluated on real robots strictly as they are, delegating the lower level continuous controls to preserve the discrete simulated action-space. Compared to existing reports, e.g.~\cite{Kadian_2020}, we did not produce common laboratory conditions --- our agents had to fight with challenging settings like sensor failures due to glass fronts and mirrors, wheels sliding on slippery floors or bumping into thick carpets.

Our main objectives were the study of internal visual reasoning learned and performed by the agent for this task, which requires to attain a target position given a (noisy) GPS signal and RGB-D observations. The agent can not blindly follow the direction indicated by the GPS signal and needs to use the RGB-D observations to estimate the geodesic path, which can significantly differ from the Euclidean path because of walls, narrow corridors and other obstacles.

\myParagraph{Classical planning vs. ML}
while we also evaluate performance of classical planners on this task for comparison, the goal of our work is \emph{not} to claim superiority of end-to-end training over the classics.  
We currently know that the PointGoal task can be addressed, ``solved'' with sufficient robustness by using symbolic planners, but they have a hard time to scale up to complex visual reasoning. The main question we explore here is whether end-to-end learning can work in general. In cases where end-to-end training is required, do we need to combine ML methods with classical low level planning when we target real physical environments, or can end-to-end training perform in its own right? In what follows we will show, that low-level waypoint navigation in physical environments can indeed be addressed by a trained planner.

In our study, we try to answer the following questions:
\begin{itemize}
\item Is PointGoal navigation in real environments possible without any adaptation from simulated training environments?
\item How are the different sensors used? When does the agent use visual observations and how? What are the regions attended to by the agent and in which situation?
\item How do the trained agents generalize to unseen conditions? Is successful navigation conditioned on memorization on a trajectory level, or do the agents learn some form of spatial reasoning transferable from simulation to the real environment?
\end{itemize}
After a description of the related work (Section \ref{sec:sota} on robot navigation and sim2real transfer), the following sections will describe the experimental setup (environments, agents, and evaluation protocol, Section \ref{sec:setup}), followed by a detailed analysis of the performance of the agents in real and simulated environments and visulizations of their sensor usage (Section \ref{sec:experiments}).

\section{Related work}
\label{sec:sota}

\myParagraph{Simulators and indoor navigation.} 
Research on robot-inspired agents that perceive, navigate, and
interact with their environment is currently carried out in simulation rather than in real physical environments, at least when based on machine learning~\cite{Savva_2019_ICCV,xiazamirhe2018gibsonenv,ai2thor}. Simulators can run experiments much faster than real-time due to high parallelization, and can deliver {\it decades} of simulated agent experience to in only {\it days} of wall-clock time~\cite{gupta17cognitive}. Moreover, evaluating agents in simulation is safer and cheaper and allows for easier automatic benchmarking of new techniques compared to handling physical robots in the real-world~\cite{choi21use}.

Simulators serve as a testbed for developing increasingly realistic indoor environments~\cite{Savva_2019_ICCV,ai2thor,igibson20} and navigational tasks~\cite{gupta17cognitive,anderson18vision}, as well as running navigation challenges~\cite{challenge20,challenge21}. 
Massive synthetic data in simulators enables learning perception and control policies in end-to-end mode\cite{gupta17cognitive}.

\myParagraph{Simulation-to-real gap.}
No simulation can perfectly replicate reality. Despite the tremendous progress in computer graphics and game engine technology, highly used in robot learning, the sim2real gap can compromise any strong performance achieved in simulation when agents are tested in the real-world. 

Perception and control policies learned in simulation often do not generalize well to real robots due to  inaccuracies in modelling, simplifications and biases. To close the gap, domain randomization methods~\cite{tan18real,peng18sim} assume that the real world distribution is a randomized instance of the simulation environment; they treat the discrepancy between the domains as variability in the simulation parameters. 

Alternatively, domain adaptation methods learn an invariant mapping function for matching distributions between the simulator and the robot environment. Related examples to the work presented here include 
sim2real transfer of visuo-motor policies for goal-directed reaching movement by adversarial learning~\cite{zang19adversarial}, adapting dynamics in reinforcement learning~\cite{eysenbach21off}, and adapting object recognition model in new domains~\cite{zhu19adapting}. 
Kadian~\etal~\cite{Kadian_2020} investigated the sim2real predictivity of Habitat-Sim~\cite{Savva_2019_ICCV} for PointGoal navigation and proposed a new metric to quantify predictivity, called Sim-vs-Real Correlation Coefﬁcient (SRCC).
Recently, Chattopadhyay~\etal~\cite{chattopadhyay2021robustnav} benchmarked the robustness of embodied navigation agents to visual and dynamics corruptions. They found that navigation agents trained in simulation severely under-perform when evaluated in corrupted target environments. The analysis of mistakes made by embodied navigation agents when operating under such corruptions has shown that although standard data-augmentation techniques and self-supervised adaptation strategies offer some improvement, much remains done in terms of fully recovering lost performance.

\myParagraph{Interpretability in RL and Computer Vision.}
Understanding and interpreting deep neural networks is an ongoing effort and research domain, and work has targeted different applications and tasks, including image classification \cite{zeiler2013visualizing, gradcam20}, but also gathering tasks in Video games simulators \cite{jaunet2020drlviz}. The latter is close to robotics, sharing navigation components and even the general meta-architecture of the agent with a recurrent memory. The authors developed a data visualization interface allowing to inspect the usage of sensors and recurrent memory, helping to understand the decision making process of the  agent. 

\begin{figure}[t]
    \centering
    \includegraphics[width=\linewidth]{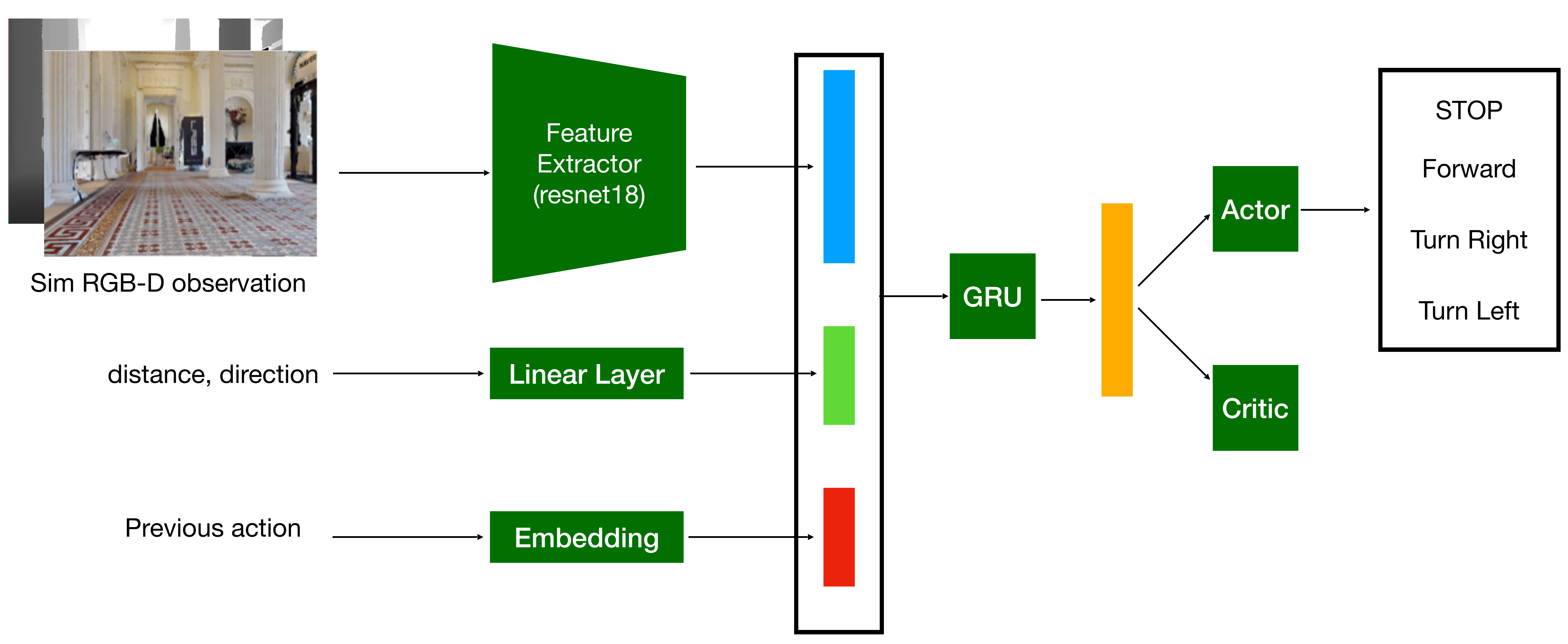}
    \caption{The architecture of the baseline RL-agent trained with PPO, taking as input visual observations, a noisy GPS signal, and the previous action.}
    \label{fig:rlagent}
\end{figure}

\begin{figure*}[t]
\centering
\captionsetup[subfigure]{justification=centering}
\begin{subfigure}[b]{.28\linewidth}
\includegraphics[width=\linewidth]{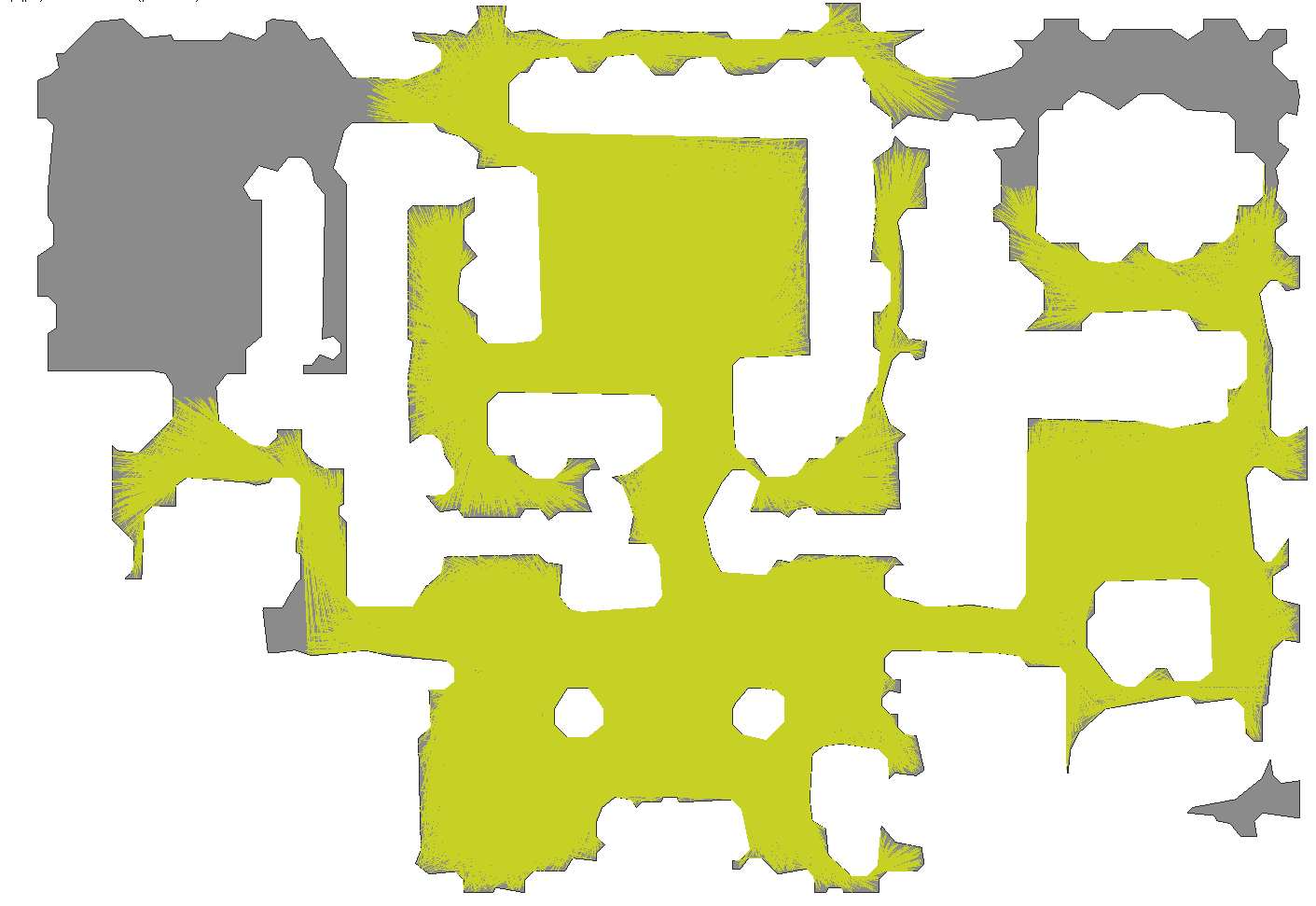}
\caption{\emph{NLE} - \\ 50,000 training episodes }\label{fig:nle:train:episodes}
\end{subfigure}
\begin{subfigure}[b]{.28\linewidth}
\includegraphics[width=\linewidth]{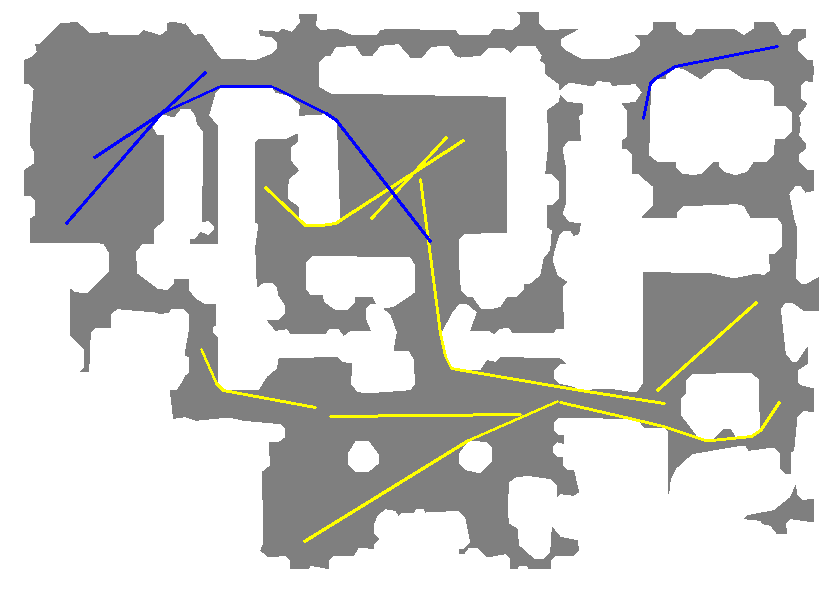}
\caption{\emph{NLE} - \\ 11 evaluation episodes}\label{fig:nle:sim2real:perfect}
\end{subfigure}
\begin{subfigure}[b]{.21\linewidth}
\includegraphics[width=\linewidth]{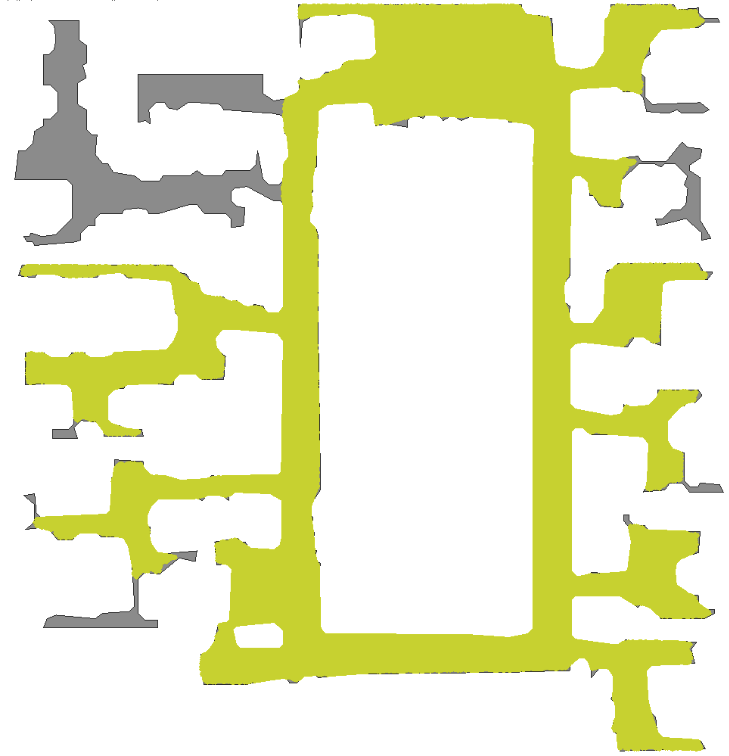}
\caption{\emph{INSAL} - \\ 50,000 training episodes}\label{fig:insal:train:episodes}
\end{subfigure}
\begin{subfigure}[b]{.21\linewidth}
\includegraphics[width=\linewidth]{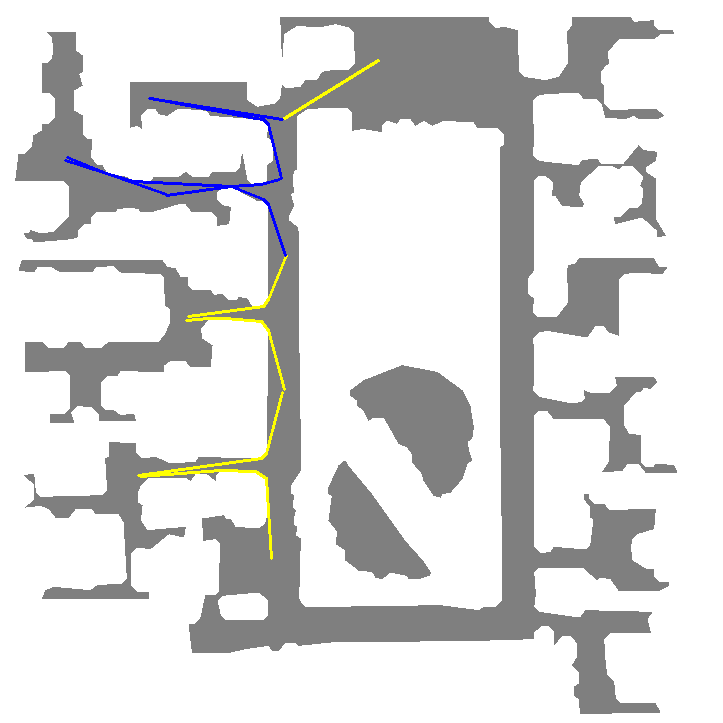}
\caption{\emph{INSAL} - \\ 10 evaluation episodes}
\label{fig:insal:sim2real:perfect}
\end{subfigure}
\caption{Overview of the 50,000 training episodes used to train the \emph{Targeted} and \emph{Finetuned} agents for both the \emph{NLE} (\ref{fig:nle:train:episodes}) and the \emph{INSAL} (\ref{fig:insal:train:episodes}) scenes. Parts of the scenes (appearing in gray) where kept out of the training set, in order to have ``seen'' ({\color{yellow!70!black} yellow}) and ``unseen'' ({\color{blue} blue}) episodes in our evaluation sets (\ref{fig:nle:sim2real:perfect}, \ref{fig:insal:sim2real:perfect}).
Note that for the \emph{Gibson} agent, \textbf{all} episodes are unseen.}
\label{fig:distribution}
\end{figure*}

\section{Experimental setup}
\label{sec:setup}

\noindent
In our experiments we train a neural policy, which is able to act in two different environments: 
\begin{itemize}
    \item a virtual environment through the \emph{Habitat} simulator~\cite{Savva_2019_ICCV} with discrete actions (\texttt{MOVE\_FORWARD}, \texttt{TURN\_LEFT}, \texttt{TURN\_RIGHT}) and \texttt{STOP}), and
    \item a real physical robot, a \emph{LoCoBot}~\cite{locobot} 
    [ \includegraphics[height=4mm]{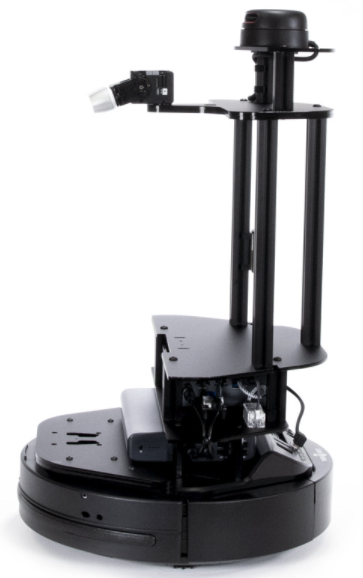} ].
\end{itemize}
All training is done in the simulated environment only, we deploy the trained agent directly to the physical robot \emph{without any adaptation in perception}. The translation of discrete actions to the continuous actions of the \emph{LoCoBot} is performed by the standard \texttt{move\_base} navigation stack of \texttt{ROS}, which also handles the navigation towards the starting position of each episode, and the evaluation of geodesic distances. 
To this end, it builds on a position estimation provided by the default \texttt{ROS} implementation of the Adaptive Monte-Carlo Localization algorithm~\cite{TBF2002probabilisticrobotics}. Instead of creating a map dynamically with SLAM, we export a map from the \emph{Habitat} simulator to share the coordinate system, which allows us to execute of the same episode in the simulator and in the real world, thus ensuring a fair comparison.

The neural RL-agent policy architecture is based on a ResNet module for perception and recurrent GRU memory, as shown in Fig.~\ref{fig:rlagent}. This agent, trained with the PPO algorithm, is the base implementation provided in the \emph{Habitat} simulator~\cite{Savva_2019_ICCV}.  
At each step, the agent receives a 160$\times$120 pixels RGB-D observation, matching the extrinsic and intrinsic parameters of the \emph{Intel RealSense} camera installed on the \emph{LoCoBot}, in particular, its position, the field of view and aspect ratio. It also receives a GPS vector (Euclidean distance and direction), describing the position of the goal relative to its current position, provided by the robot's position estimation system.

\myParagraph{Generalization over ... what?} the PointGoal task requires an agent 
to navigation from a start position 
to a target position, receiving at each instant a 
a GPS direction pointing towards the goal; in the real environment this direction is estimated by the positioning system and is often noisy. In general, naively following the GPS direction does not solve the task, as the presence of walls and obstacles requires the agent to avoid them and to find the shortest {\it geodesic} path, using the RGB-D input in addition to the GPS.

We evaluate two different generalization scenarios, \emph{both} requiring to generalize from simulation to the real world:
\begin{itemize}
    \item[\ding{192}] \textbf{Generalization to unseen environments} --- an agent is trained on 
    on a set of environments and at test time needs to navigate in an unknown environment. This corresponds to a situation where a robot is
    operating ``out of the box'', with no possibility to retrain.
    \item[\ding{193}] \textbf{Generalization to unseen trajectories} --- an agent is trained on 
    a set of environments and at test time needs to navigate in a physical environment it has been trained on, i.e. in its simulated variant. It will not necessarily navigate the same trajectories, but eventually similar ones --- we push this further by excluding regions from the environment from training (``unseen'').
    
    This use case is commercially feasible, but requires more effort at deployment, as the agent needs to be retrained on a virtual environment created from a 3D scan of the target building.
   
\end{itemize}
We performed experiments in two different physical environments: the \emph{NLE} building, a French classical castle 
and \emph{INSAL}, modern office space (Fig.~\ref{fig:teaser}). Both buildings were 3D scanned with a professional Matterport system~\cite{matterport} and loaded into the \emph{Habitat} simulator, resulting in two 
novel simulated environments suitable for training navigation agents. We then generated 50,000 training episodes for both scenes, controlling for the distribution of episode difficulties by constraining the geodesic distances between start and goal and the ratio of Euclidean to geodesic distances. 
In both scenes, we isolate an ``unseen'' region, which is excluded from the training set and reserved for evaluation (cf. Fig.~\ref{fig:distribution}). This split between ``seen'' and ``unseen'' regions is meaningless in the case when the whole test environment is excluded from training (case \ding{193} above), since then all evaluation episodes are unseen.

Eleven evaluation episodes in \emph{NLE} and ten episodes in \emph{INSAL} cross both ``seen'' and ``unseen'' regions of the scenes. It is important to note that both 
\emph{INSAL} and \emph{NLE} scenes underwent changes between the digitization and the agents evaluation phase. Therefore, even the ``seen'' parts of the scenes contain some discrepancies between simulation and real, such as moved pieces of furniture or absent/new objects. This contributes to the \emph{sim2real} gap, alongside the rendering difference between real sensors and simulated ones.

Additionally, for certain agents we also use the standard \emph{Gibson} dataset~\cite{xiazamirhe2018gibsonenv} for pre-training, which contains 3,600,000 episodes over 72 different scenes. 

We deploy three variants of the ResNet+GRU agent: 
\begin{itemize}
  \item The \colorbox{blue!25}{\emph{Gibson}} agent, trained solely on \emph{Gibson} scenes (generalization case \ding{192}).
  \item The \colorbox{red!25}{\emph{Targeted}} agents (one per scene), trained on the seen region of their respective scenes (generalization case \ding{193}).
  \item The \colorbox{green!25}{\emph{Finetuned}} agents (one per scene), first pre-trained on the \emph{Gibson} dataset, then fine-tuned to their respective scenes (generalization case \ding{193}). 
\end{itemize}
As a baseline, we also provide an evaluation of the standard \colorbox{orange!25}{\texttt{ROS}} navigation stack using the same metrics.

\myParagraph{Evaluation metrics} We used three standard metrics from the field:
\begin{itemize}
    \item {\it Success Rate} (SR) is the percentage of the episodes where the agent called \texttt{STOP} with the final distance $d_k$ to the goal being lower than a predefined threshold $d_\text{success}$:
    \begin{equation}
        SR = \frac{1}{N}\sum_{k=1}^N\mathbbm{1}_{d_k \leq d_\text{success}}
    \end{equation}
    \item {\it Success weighted by Path Length} (SPL) is the SR with each success weighted by the ratio between the optimal path length $l_k^*$ and the distance travelled by the agent $l_k$:
    \begin{equation}
        SPL = \frac{1}{N}\sum_{k=1}^N \mathbbm{1}_{d_k \leq d_\text{success}}\frac{l_k^*}{\max\{l_k,l_k^*\}}
    \end{equation}
    \item {\it Soft SPL} (sSPL) is a softer version of SPL where the Boolean success value is replaced by a continuous measure of the progress made towards the goal:
    \begin{equation}
        sSPL = \frac{1}{N}\sum_{k=1}^N \max\left\{0, 1 - \frac{d_k}{l_k^*}\right\} \frac{l_k^*}{\max\{l_k,l_k^*\}}.
    \end{equation}
\end{itemize}

\begin{table}[t]
\setlength{\tabcolsep}{2pt}
\newcommand{\bs}[1]{{\bf #1}}                                   
\newcommand{\na}[1]{{\color{black!75}\itshape\footnotesize #1}} 
\newcommand{\ccg}{\cellcolor{blue!25}}                          
\newcommand{\cct}{\cellcolor{red!25}}                           
\newcommand{\ccf}{\cellcolor{green!25}}                         
\resizebox{\columnwidth}{!}{%
\begin{tabular}{c l c c c c c c c c c c}
    \hline
                            &                       &                     &\multicolumn{3}{c}{Seen}                    &\multicolumn{3}{c}{Unseen}                  &\multicolumn{3}{c}{Overall}                 \\
    \multicolumn{2}{c}{Scene}                       &Agent                &SR            &SPL           &sSPL          &SR            &SPL           &sSPL          &SR            &SPL           &sSPL          \\
    \hline
    \hline
                            &                       &\ccg\emph{Gibson}    &\ccg\na{75.0} &\ccg\na{60.3} &\ccg\na{65.1} &\ccg    66.7  &\ccg    53.2  &\ccg    52.5  &\ccg    72.7  &\ccg    58.3  &\ccg    61.7  \\
                            &                       &\cct\emph{Targeted}  &\cct\bs{100 } &\cct\bs{94.4} &\cct\bs{93.2} &\cct\bs{100 } &\cct    72.9  &\cct    72.2  &\cct\bs{100 } &\cct    88.6  &\cct    87.5  \\
                            &\multirow{-3}{*}{sim } &\ccf\emph{Finetuned} &\ccf\bs{100 } &\ccf    93.8  &\ccf    92.7  &\ccf\bs{100 } &\ccf\bs{83.7} &\ccf\bs{82.8} &\ccf\bs{100 } &\ccf\bs{91.1} &\ccf\bs{90.0} \\
    \hhline{~-----------}
                            &                       &\ccg\emph{Gibson}    &\ccg\na{100 } &\ccg\na{73.0} &\ccg\na{71.1} &\ccg\bs{100 } &\ccg\bs{85.0} &\ccg\bs{84.3} &\ccg\bs{100 } &\ccg    76.2  &\ccg    74.7  \\
                            &                       &\cct\emph{Targeted}  &\cct    87.5  &\cct    61.9  &\cct    62.5  &\cct    33.3  &\cct    22.7  &\cct    26.8  &\cct    72.7  &\cct    51.2  &\cct    52.7  \\
    \multirow{-6}{*}{NLE}   &\multirow{-3}{*}{real} &\ccf\emph{Finetuned} &\ccf\bs{100 } &\ccf\bs{88.7} &\ccf\bs{86.3} &\ccf    66.7  &\ccf    55.6  &\ccf    54.6  &\ccf    90.9  &\ccf\bs{79.7} &\ccf\bs{77.6} \\
    \hline
    \hline
                            &                       &\ccg\emph{Gibson}    &\ccg\na{100 } &\ccg\na{89.4} &\ccg\na{87.9} &\ccg\bs{100 } &\ccg    78.4  &\ccg    76.5  &\ccg\bs{100 } &\ccg    83.9  &\ccg    82.2  \\
                            &                       &\cct\emph{Targeted}  &\cct\bs{100 } &\cct    91.3  &\cct    88.8  &\cct    40.0  &\cct    26.0  &\cct    26.0  &\cct    70.0  &\cct    58.6  &\cct    67.0  \\
                            &\multirow{-3}{*}{sim } &\ccf\emph{Finetuned} &\ccf\bs{100 } &\ccf\bs{93.6} &\ccf\bs{92.2} &\ccf\bs{100 } &\ccf\bs{94.5} &\ccf\bs{92.2} &\ccf\bs{100 } &\ccf\bs{94.1} &\ccf\bs{92.2} \\
    \hhline{~-----------}
                            &                       &\ccg\emph{Gibson}    &\ccg\na{20.0} &\ccg\na{16.9} &\ccg\na{16.9} &\ccg    80.0  &\ccg    46.4  &\ccg    45.4  &\ccg    50.0  &\ccg    31.7  &\ccg    31.2  \\
                            &                       &\cct\emph{Targeted}  &\cct    80.0  &\cct    43.6  &\cct    42.5  &\cct    80.0  &\cct    50.6  &\cct    49.4  &\cct    80.0  &\cct    47.1  &\cct    45.9  \\
    \multirow{-6}{*}{INSAL} &\multirow{-3}{*}{real} &\ccf\emph{Finetuned} &\ccf\bs{100 } &\ccf\bs{92.8} &\ccf\bs{90.6} &\ccf\bs{100 } &\ccf\bs{91.6} &\ccf\bs{89.1} &\ccf\bs{100 } &\ccf\bs{92.2} &\ccf\bs{89.8} \\
    \hline
\end{tabular}%
}
\caption{Quantitative results for experiments in simulation and real physical robots. We report the average SR, SPL and sSPL on seen, unseen and all episodes. \bs{Bold} font highlights best values on the corresponding set of episodes. \na{Gray} font denotes that the \emph{Gibson} agent has not really seen any part of the scenes during training.}
\label{tab:expe:sim2real}
\end{table}

\myParagraph{Implementation} we developed our own interface, publicly available online\footnote{\url{https://github.com/wgw101/habitat_sim2real}}, which makes the real robot and its entire \texttt{ROS} stack appear as a ``Simulator'' in the \texttt{habitat-lab} framework. This allows to transparently switch from a simulation evaluation to real world one by changing a single line in the configuration file. This allows us to re-use all the metrics and features already present in \texttt{habitat-lab} for the \emph{PointNav} task, and to deploy the same agent in simulation and on the \emph{LoCoBot}.
We rely on \texttt{ROS} to provide the position estimation we need, which affects the GPS vector given to the agent, but more importantly the precision of our evaluation metrics.

\section{Experimental Results}
\label{sec:experiments}

\myParagraph{Quantitative performance analysis} 
Table~\ref{tab:expe:sim2real} shows the agent performance 
on real robot and in simulation, in both environments. We observe
a performance drop between simulation and real, but also that this gap is quite low for the Finetuned agent. We conjecture that pre-training on a large number of additional simulated environments 
drastically reduces the sim2real gap.

Complete generalization to unseen environments (generalization case \ding{192}) is possible but not universal, as shows the poor performance of the Gibson agent in the \emph{INSAL} environment. Deployment of a robot to real environment after pre-training on a 3D scanned variant (case \ding{193}), however, leads to surprisingly good performance. For the Targeted and Finetuned agents (case \ding{193}), the performance gap between seen and unseen episodes is low --- training on an environment benefits performance even on the unseen parts, which, unsurprisingly, suggests the existence of factors of variation common to the trajectories of a given environment (and thus particular to the environment).

\begin{figure*}[t]
\centering
\begin{subfigure}[b]{.17\linewidth}
\includegraphics[width=\linewidth]{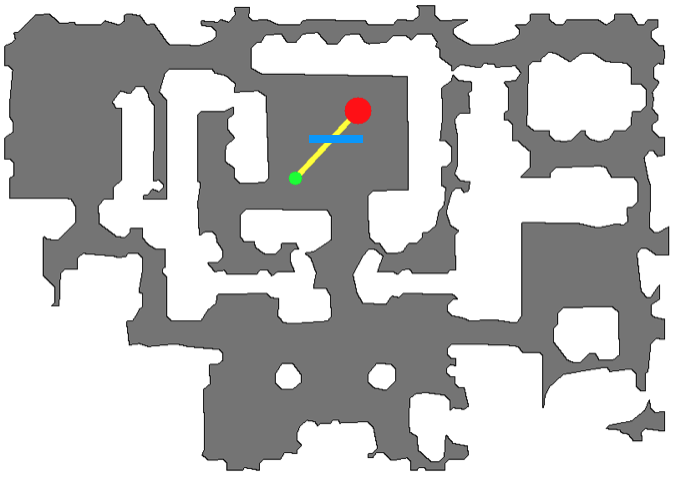}
\caption{Episode}\label{fig:episode_0_obstacles}
\end{subfigure}
\begin{subfigure}[b]{.17\linewidth}
\includegraphics[width=\linewidth]{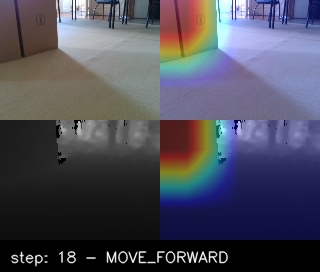}
\caption{Step 18}
\end{subfigure}
\begin{subfigure}[b]{.17\linewidth}
\includegraphics[width=\linewidth]{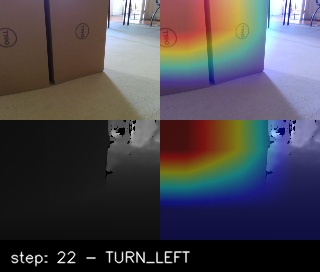}
\caption{Step 22}
\end{subfigure}
\begin{subfigure}[b]{.17\linewidth}
\includegraphics[width=\linewidth]{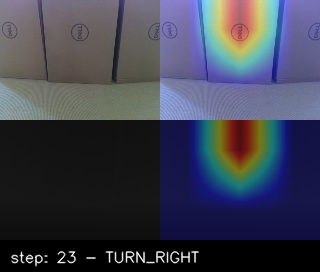}
\caption{Step 23}
\end{subfigure}
\begin{subfigure}[b]{.17\linewidth}          
\includegraphics[width=\linewidth]{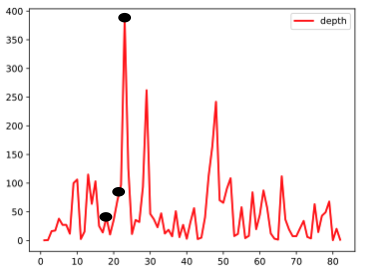}
\caption{Depth Saliency}
\end{subfigure}

\begin{subfigure}[b]{.17\linewidth}
\includegraphics[width=\linewidth]{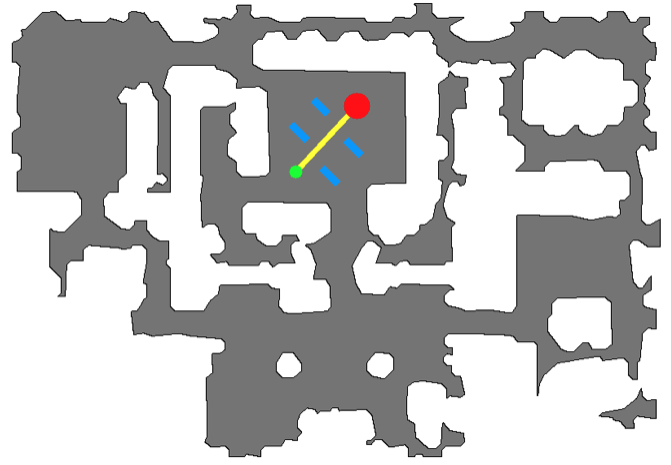}
\caption{Episode}
\end{subfigure}
\begin{subfigure}[b]{.17\linewidth}
\includegraphics[width=\linewidth]{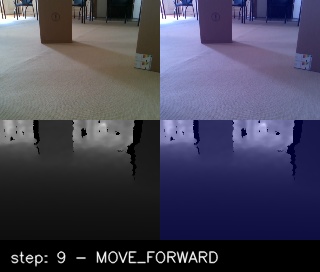}
\caption{step 9}
\end{subfigure}
\begin{subfigure}[b]{.17\linewidth}
\includegraphics[width=\linewidth]{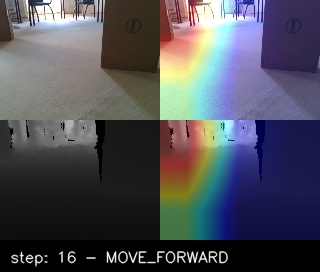}
\caption{step 16}
\end{subfigure}
\begin{subfigure}[b]{.17\linewidth}
\includegraphics[width=\linewidth]{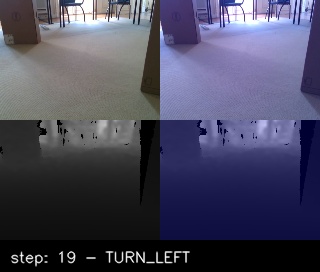}
\caption{step 19}
\end{subfigure}
\begin{subfigure}[b]{.17\linewidth}          
\includegraphics[width=\linewidth]{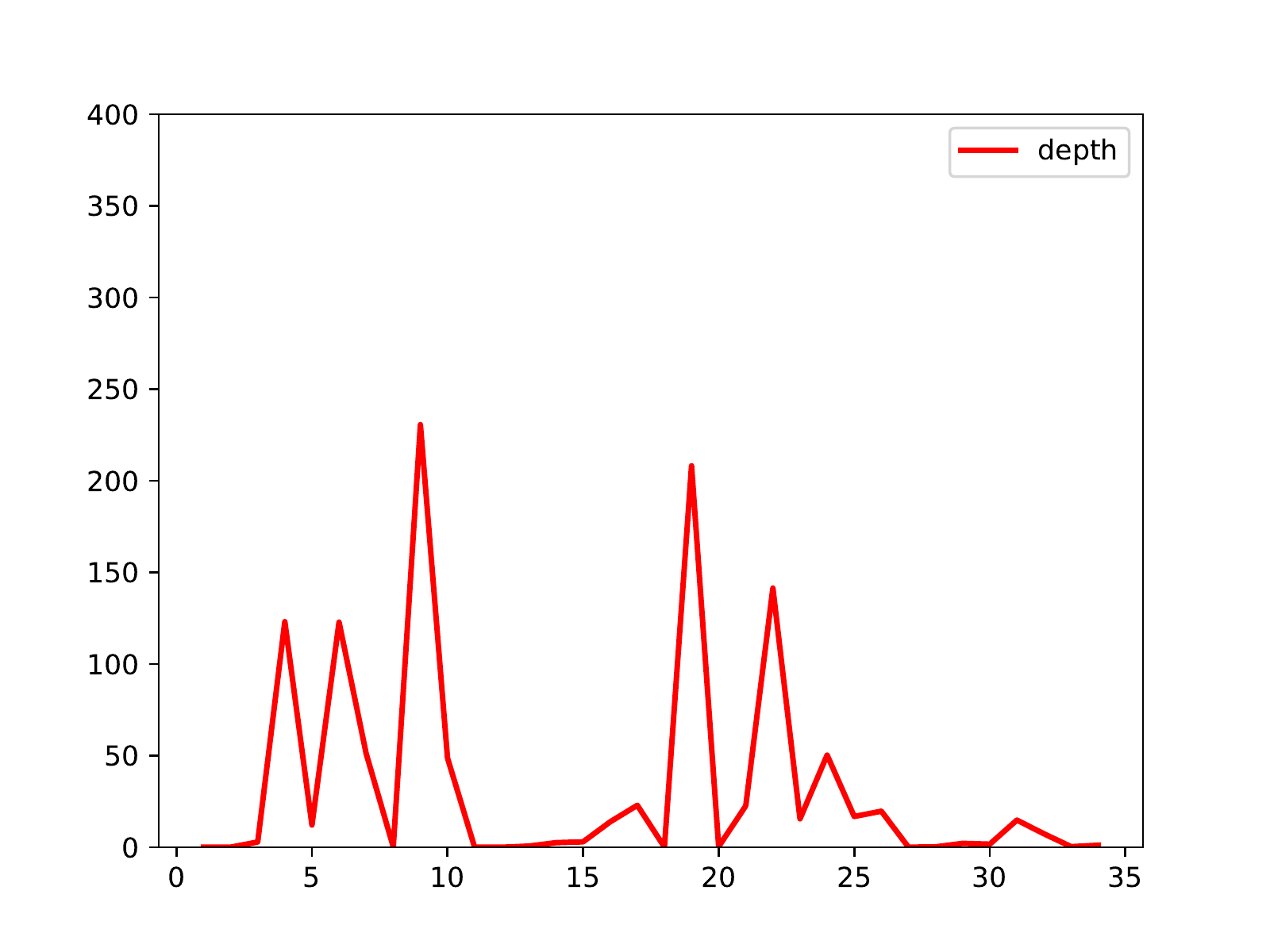}
\caption{Depth Saliency}
\end{subfigure}
\caption{Impact of obstacles on visual attention. Top: obstacles (boxes) are close and on the path to target, the robot 
pays the main attention to them. Bottom: the same boxes are visible in the scene, but 
do not block the navigation towards target.
Black dots on the saliency curve indicate the steps shown.}
\label{fig:episode_0}
\end{figure*}

\begin{figure*}[t] \centering
\begin{subfigure}[b]{.17\linewidth}
\includegraphics[width=\linewidth]{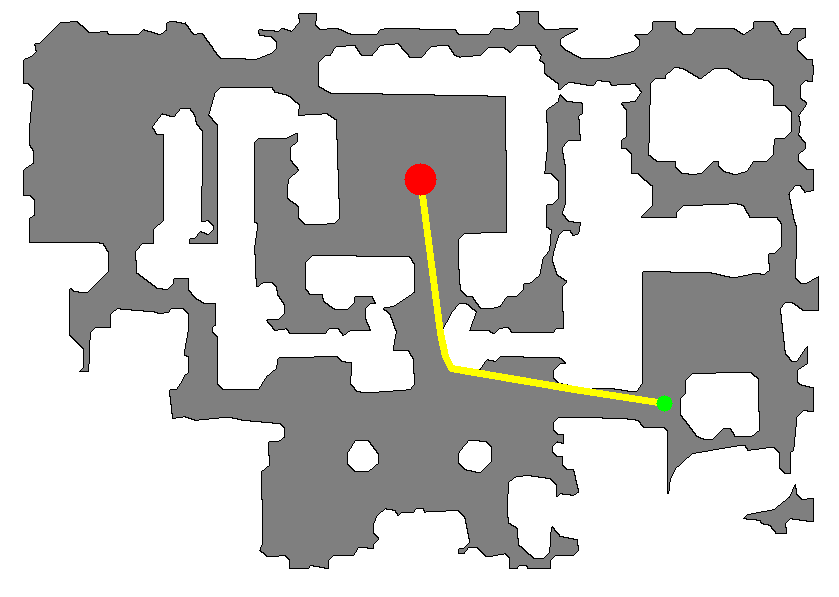}
\caption{Episode}
\end{subfigure}
\begin{subfigure}[b]{.17\linewidth}
\includegraphics[width=\linewidth]{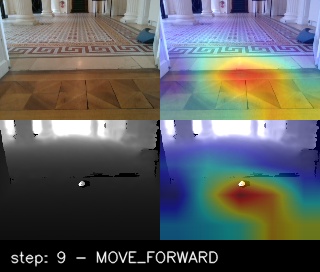}
\caption{step 9}
\end{subfigure}
\begin{subfigure}[b]{.17\linewidth}
\includegraphics[width=\linewidth]{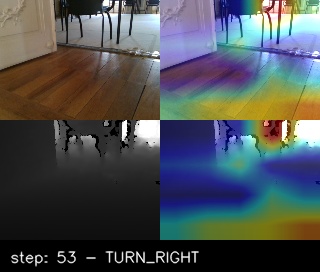}
\caption{step 53}
\end{subfigure}
\begin{subfigure}[b]{.17\linewidth}          
\includegraphics[width=\linewidth]{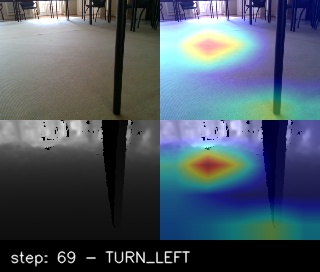}
\caption{step 69}
\end{subfigure}
\begin{subfigure}[b]{.17\linewidth}
\includegraphics[width=\linewidth]{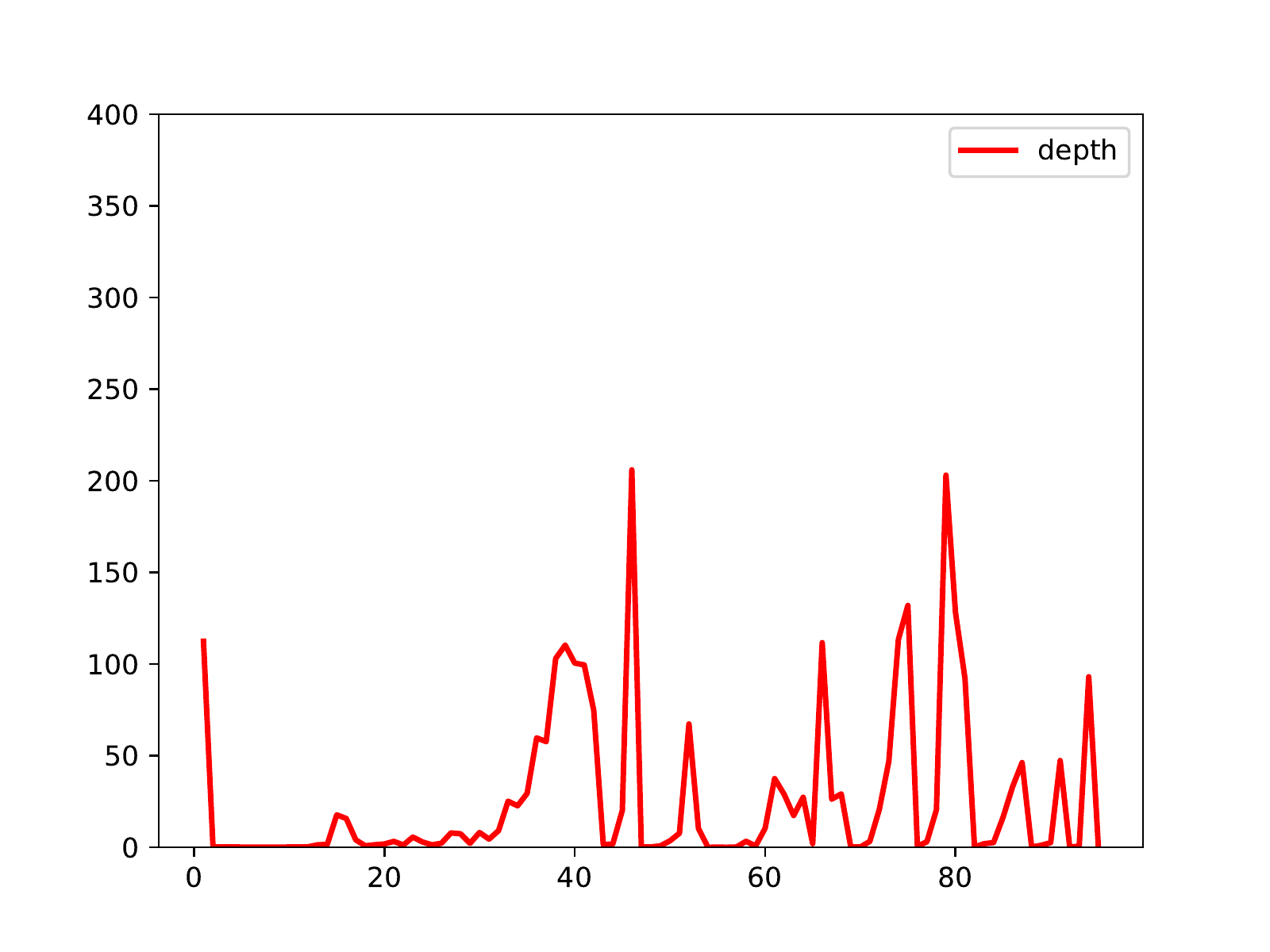}
\caption{Depth Saliency}
\end{subfigure}
\caption{Impact of navigation in a non-straight L-shape line on visual attention. The agent 
pays attention to visual input when it considers a direction change.}
\label{fig:episode_long_L_shape}
\end{figure*}

\begin{figure*}[t] \centering
\begin{subfigure}[b]{.15\linewidth}
\includegraphics[width=\linewidth]{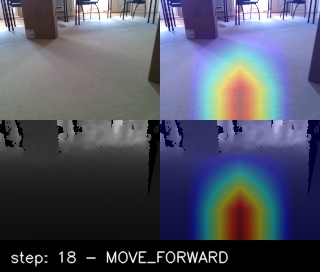}
\caption{\texttt{MOVE\_FORWARD}}\label{fig:A}
\end{subfigure}
\begin{subfigure}[b]{.15\linewidth}
\includegraphics[width=\linewidth]{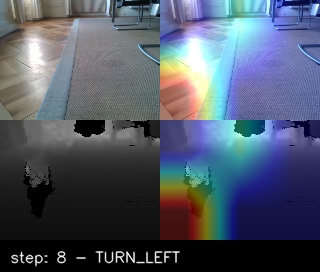}
\caption{\texttt{TURN\_LEFT}}\label{fig:B}
\end{subfigure}
\begin{subfigure}[b]{.15\linewidth}
\includegraphics[width=\linewidth]{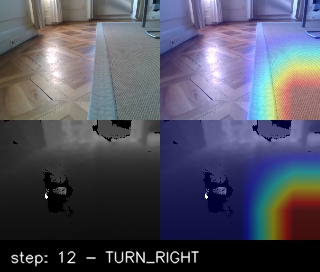}
\caption{\texttt{TURN\_RIGHT}}\label{fig:C}
\end{subfigure}
\begin{subfigure}[b]{.15\linewidth}
\includegraphics[width=\linewidth]{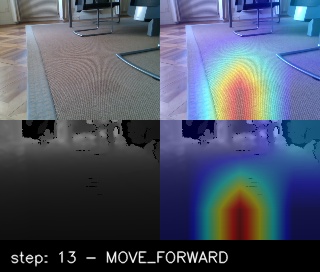}
\caption{\texttt{MOVE\_FORWARD}}\label{fig:D}
\end{subfigure}
\begin{subfigure}[b]{.15\linewidth}          
\includegraphics[width=\linewidth]{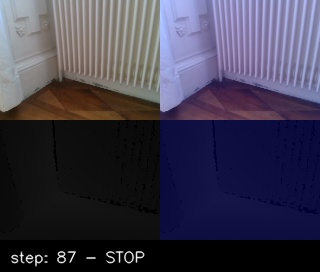}
\caption{\texttt{STOP}}\label{fig:E}
\end{subfigure}
\begin{subfigure}[b]{.15\linewidth}          
\includegraphics[width=\linewidth]{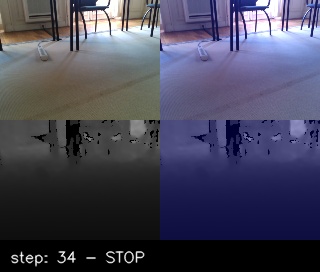}
\caption{\texttt{STOP}}\label{fig:F}
\end{subfigure}
\caption{Visual attention and motion: the agent has a strong tendency to attend to regions it will navigate to through turns.}
\label{fig:gradcam_different_actions}
\end{figure*}

\myParagraph{Visualization and interpretation}
We have explored different ways to increase interpretability of the neural model and to visualize parts of its sensor usage. We ask the following questions: (a) what type of sensor is important at what time in an episode (RGB-D or GPS), (b) which region is attended to in an observed image, and ultimately, (c) why does an agent take a certain action at a certain moment?

We use the visualization procedure grad-CAM~\cite{gradcam20}, which computes a measure of attention or importance, by, written in simplified terms, calculating the gradients of the output layer of the policy corresponding to the taken action w.r.t. to the inputs or an intermediate neural unit. A high derivative for a pixel, value or unit value suggests a high impact on the decision. We chose the last convolutional layer in the feature extractor of the RGB-D observation and overlay its gradient on the input image in pseudo colors, c.f.  Figs.~\ref{fig:episode_0}, \ref{fig:episode_long_L_shape} and \ref{fig:gradcam_different_actions}. In these figures we also provide an indication of the importance of of the visual input, 
which corresponds to the accumulated gradients. We saw that the importance of the GPS sensor stayed relatively flat.

In Figure \ref{fig:episode_0} (top), we visualize the impact of blocking obstacles (boxes) placed in front of the goal, on sensor usage. This obstacle should in principle require the agent to use the visual observation, as the GPS direction does not point towards a feasible path. The agent indeed focuses on the obstacles as they appear in front of the camera, verified in the Grad-CAM heatmap. In the  selected three steps shown in the Figure we can see that, when the agent was fully blocked by the obstacles, the saliency value of the depth image peaks, indicating its influence on the chosen avoidance actions: \texttt{TURN\_LEFT} in step 22 and  \texttt{TURN\_RIGHT} in step 23 as counter-action. After the peak is reached, the agent had to navigate a half-circle around the obstacle, during which the importance of the depth input started to decreases until the goal was reached. To check whether attention is caused by visual saliency of the cardboard boxes, we created a ``control episode'' where these boxes are in a non-blocking placement --- the agent had a significantly decreased (or no) focus on the depth, except when it needed to verify that blockage might happen, in step 16.

In Figure \ref{fig:episode_long_L_shape} we visualize an episode with an L-shaped trajectory, thus different from a straight path, which would have otherwise allowed to plainly follow the GPS directions. The depth input was ignored in the first part of the episode consisting mostly of forwards moves. However, once the agent needs to find the turning point, requiring to differentiate between \texttt{MOVE\_FORWARD} and \texttt{TURN\_RIGHT}, depth usage jumps up and the heatmap indicates that the action taken at each time step is strongly related to the regions where the agent looks at (left region for left turns etc.). We have cross checked several episodes to confirm this tendency, shown in Figure \ref{fig:gradcam_different_actions} --- the agent has a strong tendency to look at the region towards which it will navigate with a turning action. Not surprisingly, the visual input is unused when the goal is reached and \texttt{STOP} is called, as this can be decided from the distance value.

\begin{table}[t]
\setlength{\tabcolsep}{2pt}
\newcommand{\bs}[1]{{\bf #1}}                                   
\newcommand{\na}[1]{{\color{black!75}\itshape\footnotesize #1}} 
\newcommand{\ccf}{\cellcolor{green!25}}                         
\newcommand{\ccr}{\cellcolor{orange!25}}                        
\resizebox{\columnwidth}{!}{%
\begin{tabular}{c l c c c c c c c c c c}
    \hline
                            &                       &                     &\multicolumn{3}{c}{Seen}                    &\multicolumn{3}{c}{Unseen}                  &\multicolumn{3}{c}{Overall}                 \\
    \multicolumn{2}{c}{Scene}                       &Agent                &SR            &SPL           &sSPL          &SR            &SPL           &sSPL          &SR            &SPL           &sSPL          \\
    \hline
    \hline
                            &                       &\ccf\emph{Finetuned} &\ccf\bs{100 } &\ccf\bs{88.7} &\ccf\bs{86.3} &\ccf    66.7  &\ccf    55.6  &\ccf    54.6  &\ccf    90.9  &\ccf    79.7  &\ccf    77.6  \\
    \multirow{-2}{*}{NLE}   &\multirow{-2}{*}{real} &\ccr\texttt{ROS}     &\ccr\bs{100 } &\ccr    79.7  &\ccr    78.3  &\ccr\na{100 } &\ccr\na{96.5} &\ccr\na{95.4} &\ccr\bs{100 } &\ccr\bs{84.3} &\ccr\bs{83.0} \\
    \hline
    \hline
                            &                       &\ccf\emph{Finetuned} &\ccf\bs{100 } &\ccf    92.8  &\ccf    90.6  &\ccf\bs{100 } &\ccf\bs{91.6} &\ccf    89.1  &\ccf\bs{100 } &\ccf    92.2  &\ccf    89.8  \\
    \multirow{-2}{*}{INSAL} &\multirow{-2}{*}{real} &\ccr\texttt{ROS}     &\ccr\bs{100 } &\ccr\bs{93.9} &\ccr\bs{92.8} &\ccr\na{100 } &\ccr\na{91.6} &\ccr\na{90.4} &\ccr\bs{100 } &\ccr\bs{92.8} &\ccr\bs{90.4} \\
    \hline
\end{tabular}%
}
\caption{Quantitative results for experiments on real physical robots comparing the classical planner (ROS) to the best trained agent. \na{Gray} font denotes that the \texttt{ROS} agent was given the full map of the scenes, not just the seen regions.}
\label{tab:expe:learned:vs:classical}
\end{table}

\begin{figure}[t]
    \centering
    \includegraphics[width=\linewidth]{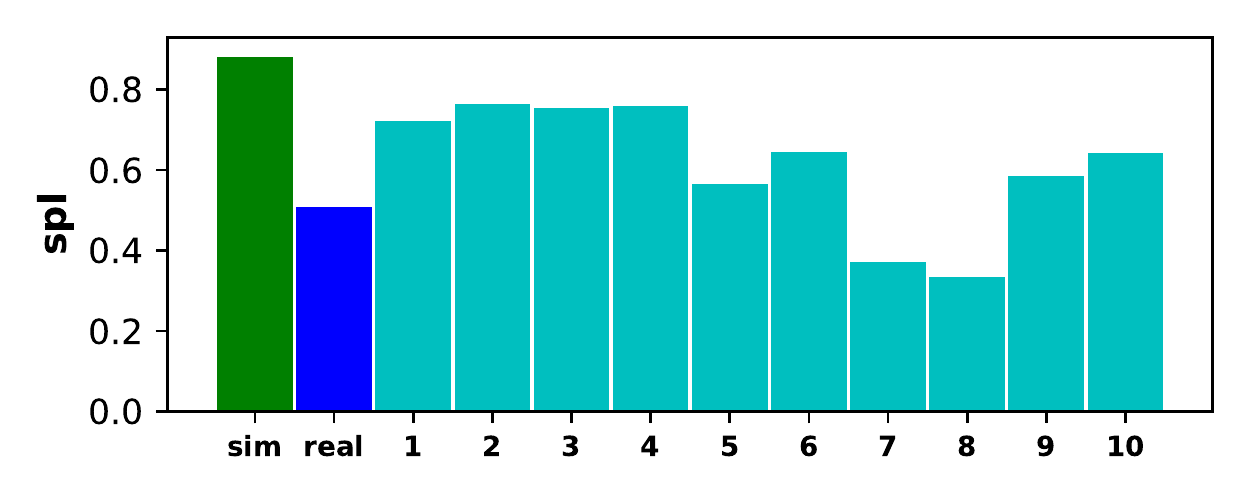}
    \caption{\label{fig:nle_envs_spl_all}Performance on noisy sim (Targeted agent), compared to sim and real (\textcolor{blue}{blue}). Noise variants are (\textcolor{cyan}{cyan}): \\
    { \footnotesize
    (1-4): have Redwood Depth noise of intensity 6 and RGB noises based on different distributions:
    1: Gaussian, 2: Speckle, 3: Salt \& Pepper, 4: Poisson; \\
    (5-6): applied Gaussian noises with different Redwood Depth Noises (intensity 3 and 9 respectively). \\
    (7-9): Gaussian noises simulated on actuators based on three common controllers described by \cite{DBLP:journals/corr/abs-1906-08236}: 7: Proportional Controller (P), 8: Dynamic Window Approach Controller from Movebase (MB), 9: Linear Quadratic Regulator (ILQR). \\
    (10): noise settings of CVPR Habitat challenge 2021 (Gaussian, Redwood with Intensity 1 and Proportional Controller noises).}}
\end{figure}

\begin{figure}[t] \centering
    \includegraphics[width=0.7 \linewidth]{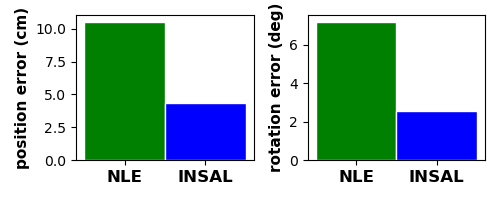}
    \caption{Error in position (left) and angle (right) between the controller target and the actual motion done by the robot, on the two scenes: ``\emph{NLE}'' (\textcolor{green}{green})  and ``\emph{INSAL}'' (\textcolor{blue}{blue}). Errors are w.r.t. to the estimate obtained by the ROS-NavStack.}\label{fig:estimated_errors}
\end{figure}

\myParagraph{Classical vs. ML} Table~\ref{tab:expe:learned:vs:classical} compares the best trained agent to the classical baselines from the ROS navigation stack. As expected, the classical baselines slightly outperform the trained agent, but their performance is quite comparable. We insist again, that this comparison was not the goal, we report it for the sake of completeness. 

\myParagraph{Comparison Real vs. Noisy Sim}
The evaluation of a physical robot in the real world is a time-consuming exercise, we therefore provide a comparison to another common practice in embodied computer vision and robotics, evaluation in noisy simulated environments. The goal is to evaluate whether ``noisy sim'' can and should be chosen as a proxy for real experiments, and which types of noise should be chosen.

We evaluated the agent on the same evaluation episodes (``\emph{NLE}'' environment) on different environment variants, which differ in noise configurations, as shown in Figure~\ref{fig:nle_envs_spl_all}. We can see that most of the noise techniques on the visual sensors are not representative and do not strongly impact performance. However, noise on the actuators does have a strong impact and can show similar or even worse performance than in the real environment. 

We investigate actuator noise further by measuring the difference between the position difference planned by a single control step and the actual position difference performed by the robot, shown in Figure~\ref{fig:estimated_errors}.
We found that the physical agents need to perform far more actions for the same trajectory than the simulated one. The  gap in error is also partly due to difficult floor conditions ---  the ``\emph{NLE}'' scene was furnished with thick carpets, which created more challenges to the robot than the more modern ``\emph{INSAL}'' scene.

\section{Conclusion}
\label{sec:conclusion}

\noindent
We have evaluated three variants of agents on two different real physical environments to benchmark the generalization capabilities of physical robots agent in the real world. We showed that for the PointGoal task, an agent pre-trained on wide range of scenes and finetuned on a targeted scene in simulation can reach a high performance and reduce the sim2real gap without the need of any sim2real transfer technique. We also conducted in depth visualization for the sensor usage in the neural network to understand the visual reasoning of the agent, showing that the agent indeed puts a attention on the visual information when needed,.


\bibliographystyle{plain} 
\bibliography{main}

\end{document}